\documentclass[10pt,twocolumn,letterpaper]{article}

\usepackage{iccv}
\usepackage{times}
\usepackage{epsfig}
\usepackage{graphicx}
\usepackage{amsmath}
\usepackage{amssymb}
\usepackage{booktabs}
\usepackage{multirow}
\usepackage{bbm}
\usepackage[caption=false]{subfig}
\usepackage{enumitem}
\usepackage{algorithmicx}
\usepackage[ruled]{algorithm2e}
\usepackage{algpseudocode}
\usepackage{appendix}

\newcommand{\frameworkName}{OnPro\xspace}
\newcommand{\methodname}{OPE\xspace}
\newcommand{\dataaugname}{APF\xspace}
\newcommand{\std}[1]{\footnotesize{#1}}

\newlength\savewidth\newcommand\shline{\noalign{\global\savewidth\arrayrulewidth\global\arrayrulewidth1.25pt}\hline\noalign{\global\arrayrulewidth\savewidth}}

\usepackage[pagebackref=true, breaklinks=true,colorlinks,citecolor=citecolor, linkcolor=linkcolor, bookmarks=false]{hyperref}
\definecolor{citecolor}{HTML}{1F801F}
\definecolor{linkcolor}{HTML}{ED1C24}
\iccvfinalcopy 

\renewcommand{\paragraph}[1]{\noindent\textbf{#1}}

\def\iccvPaperID{xxxx} 

\begin{document}

\title{Online Prototype Learning for Online Continual Learning}

\author{Yujie Wei$^{1}$\qquad Jiaxin Ye$^{1}$\qquad Zhizhong Huang$^{2}$\qquad Junping Zhang$^{2}$\qquad Hongming Shan$^{1,3,4}$\thanks{Corresponding author}
\\
$^{1}$ Institute of Science and Technology for Brain-inspired Intelligence, Fudan University\\
$^{2}$ School of Computer Science,
Fudan University\\
$^{3}$ MOE Frontiers Center for Brain Science, Fudan University\\
$^{4}$ Shanghai Center for Brain Science and Brain-inspired Technology\\
{\tt\small \{yjwei22, jxye22\}@m.fudan.edu.cn},\quad
{\tt\small \{zzhuang19, jpzhang, hmshan\}@fudan.edu.cn}
}

\maketitle
\ificcvfinal\thispagestyle{empty}\fi

\begin{abstract}
    Online continual learning (CL) studies the problem of learning continuously from a single-pass data stream while adapting to new data and mitigating catastrophic forgetting. 
    Recently, by storing a small subset of old data, replay-based methods have shown promising performance. 
    Unlike previous methods that focus on sample storage or knowledge distillation against catastrophic forgetting, this paper aims to understand why the online learning models fail to generalize well from a new perspective of shortcut learning. 
    We identify shortcut learning as the key limiting factor for online CL, where the learned features may be biased, not generalizable to new tasks, and may have an adverse impact on knowledge distillation. 
    To tackle this issue, we present the online prototype learning (\frameworkName) framework
    for online CL. 
    First, we propose online prototype
    equilibrium to learn representative features against shortcut learning and discriminative features to avoid class confusion, ultimately achieving an equilibrium status that separates all seen classes well while learning new classes.
    Second, with the feedback of online prototypes,
    we devise a novel adaptive prototypical feedback mechanism to sense the classes that are easily misclassified and then enhance their boundaries.
    Extensive experimental results on widely-used benchmark datasets demonstrate the superior performance of \frameworkName over the state-of-the-art baseline methods.
    Source code is available at \url{https://github.com/weilllllls/OnPro}.
\end{abstract}

\section{Introduction}
\label{intro}
Current artificial intelligence systems~\cite{ResNet, DNN, VGG, ViT} have shown excellent performance on the tasks at hand; however, they are prone to forget previously learned knowledge while learning new tasks, known as 
\emph{catastrophic forgetting}~\cite{catastrophic, catastrophic2, EWC}. 
Continual learning (CL)~\cite{survey1, survey2, survey3, CL_CIL1} aims to learn continuously from a non-stationary data stream while adapting to new data and mitigating catastrophic forgetting, offering a promising path to human-like artificial general intelligence.
Early CL works consider the task-incremental learning (TIL) setting, where the model selects the task-specific component for classification with task identifiers~\cite{regular1, kd1, para-iso1, survey3}. However, this setting lacks flexibility in real-world scenarios.
In this paper, we focus on a more general and realistic setting---the class-incremental learning (CIL) in the online CL mode~\cite{online_survey, onlineCL1, onlineCL2, ASER}---where the model learns incrementally classes in a sequence of tasks from a single-pass data stream and cannot access task identifiers at inference.

Various online CL methods have been proposed to mitigate catastrophic forgetting~\cite{ASER, SCR, DVC, online_pro_accum, ER, ER_AML, onlineCL1}. Among them, replay-based methods~\cite{ER,  SCR, OCM, MIR, DVC} have shown promising performance by storing a subset of data from old classes as exemplars for experience replay. 
Unlike previous methods that focus on sample storage~\cite{ASER, GSS},
we are interested in how generalizable the learned features are to new classes, and aim to understand why the online learning models fail to generalize well from a new perspective of shortcut learning.

\begin{figure}
    \centering
    \includegraphics[width=0.9\linewidth]{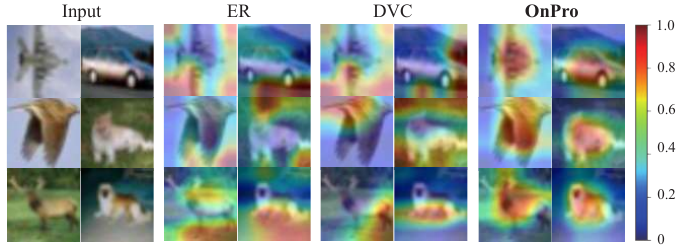}
    \caption{The visual explanations by GradCAM++ on the training set of CIFAR-10 (image size $32 \times 32$).
    Although all methods predict the correct class, shortcut learning still exists in ER and DVC.}
    \label{fig:heatmap}
\end{figure}

Intuitively, the neural network tends to ``take shortcuts''~\cite{shortcut} and focuses on simplistic features.
\textit{This behavior of shortcut learning is especially serious in online CL}, since the model may learn biased and inadequate features from the single-pass data stream.
Specifically, the model may be more inclined to learn trivial solutions \emph{unrelated} to objects, which are hard to generalize and easily forgotten. 
Take Fig.~\ref{fig:heatmap} as an example, when classifying two classes, saying airplanes in the sky and cat on the grass, the model may easily identify the shortcut clue between two classes---blue sky vs. green grass---unfortunately, the learned features are delicate and unrelated to the classes of interest. When new bird and deer classes come, which may also have sky or grass, the model has to be updated due to inapplicable previous knowledge, leading to poor generalization and catastrophic forgetting.
Thus, learning \emph{representative} features that best characterize the class is crucial to resist shortcut learning and catastrophic forgetting, especially in online CL.

In addition, 
the intuitive manifestation of catastrophic forgetting is the confusion between classes. 
To alleviate class confusion, many works~\cite{OCM, kd1, iCaRL, DER++, Co2L, protoAug} employ self-distillation~\cite{kd_work, kd_work2} to preserve previous knowledge.
However, the premise for knowledge distillation to succeed is that the model has learned sufficient discriminative features in old classes, and these features still remain discriminative when learning new classes. As mentioned above, the model may learn  oversimplified features due to shortcut learning, significantly compromising the generalization to new classes.
Thus, distilling these biased features may have an adverse impact on new classes. 
In contrast, we consider a more general paradigm to maintain discrimination among all seen classes, which can tackle the limitations of knowledge distillation.

In this paper, we aim to learn representative features of each class and discriminative features between classes, both crucial to mitigate catastrophic forgetting.
Toward this end, we present the Online Prototype learning (\frameworkName) framework for online continual learning.
The online prototype introduced is defined as ``a representative embedding for a group of instances in a mini-batch.'' There are two reasons for this design: (1) for new classes, the data arrives sequentially from a single-pass stream, and we cannot access all samples of one class at any time step (iteration); and (2) for old classes, computing the prototypes of all samples in the memory bank at each time step is computationally expensive, especially for the online scenario with limited resources. 
Thus, \emph{our online prototypes only utilize the data available at the current time step (\ie, data within a mini-batch), which is more suitable for online CL}.

To resist shortcut learning in online CL and maintain discrimination among seen classes, we first propose Online Prototype Equilibrium (\methodname) to learn representative and discriminative features for achieving an equilibrium status that separates all seen classes well while learning new classes.
Second, 
instead of employing knowledge distillation that may distill unfaithful knowledge from previous models,
we devise a novel Adaptive Prototypical Feedback (\dataaugname) 
that can leverage the feedback of online prototypes to first sense the classes---that are easily misclassified---and then adaptively enhance their decision boundaries.

The contributions are summarized as follows. 
\begin{enumerate}[leftmargin=12pt, itemsep=0pt, topsep=0pt, partopsep=0pt, noitemsep]
    \item[1)] We identify shortcut learning as the key limiting factor for online CL, where the learned features may be biased, not generalizable to new tasks, and may have an adverse impact on knowledge distillation. To the best of our knowledge, this is the first time to identify the shortcut learning issues in online CL, offering new insights into why online learning models fail to generalize well. 
    \item[2)] We present the online prototype learning framework for online CL, in which the proposed online prototype equilibrium encourages learning representative and discriminative features while adaptive prototypical feedback leverages the feedback of online prototypes to 
    sense easily misclassified classes and enhance their boundaries. 
    \item[3)] Extensive experimental results on widely-used benchmark datasets demonstrate the superior performance of our method over the state-of-the-art baseline methods.
\end{enumerate}

\begin{figure*}[htp]
  \centering
  \includegraphics[width=0.9\linewidth]{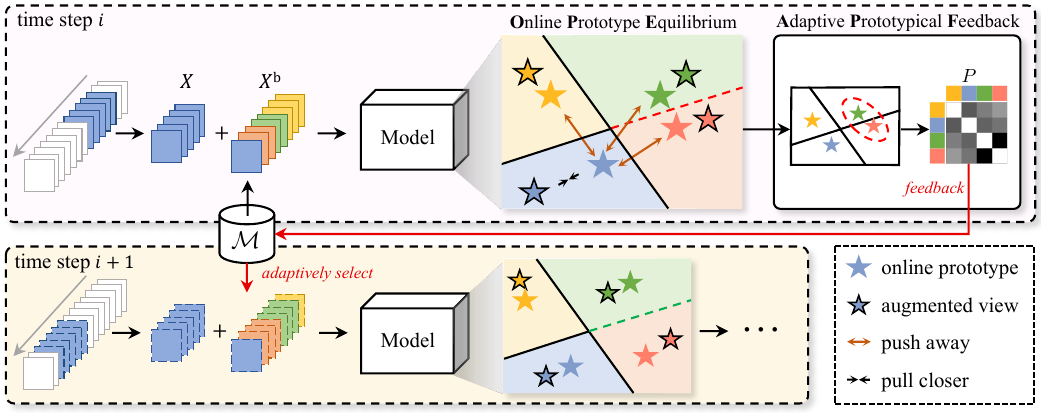}
  \caption{Illustration of the proposed \frameworkName framework. 
  At time step (iteration) $i$, the incoming data $X$ and replay data $X^\mathrm{b}$ are augmented and fed to the model to learn features with \methodname.
  Then, the proposed \dataaugname senses easily misclassified classes from all seen classes and enhances their decision boundaries.
  Concretely, \dataaugname adaptively selects more data for mixup according to the probability distribution $P$.
  }
  \label{fig:framework}
\end{figure*}

\section{Related Work}
\paragraph{Continual learning.}
Continual learning methods can be roughly summarized into three categories: regularization-based, parameter-isolation-based, and replay-based methods. Regularization-based methods~\cite{EWC, regular1, regular2, regular3} add extra regularization constraints on network parameters to mitigate forgetting. Parameter-isolation-based methods~\cite{PNN, para-iso1, para-iso2, pathNet} avoid forgetting by dynamically allocating parameters or modifying the architecture of the network. Replay-based methods~\cite{ER, MIR, GSS, AGEM, DER++, GDumb} maintain and update a memory bank (buffer) that stores exemplars of past tasks. 
Among them, replay-based methods are the most popular for their simplicity yet efficiency. Experience Replay~\cite{ER} randomly samples from the buffer. MIR~\cite{MIR} retrieves buffer samples by comparing the interference of losses. Furthermore, in the online setting, ASER~\cite{ASER} introduces a buffer management theory based on the Shapley value. SCR~\cite{SCR} utilizes supervised contrastive loss~\cite{SupCL} for training and the nearest-class-mean classifier for testing. OCM~\cite{OCM} prevents forgetting through mutual information maximization.

Unlike these methods that focus on selecting which samples to store or learning features only by instances, our work rethinks the catastrophic forgetting from a new shortcut learning perspective, and proposes to learn representative and discriminative features through online prototypes.

\paragraph{Knowledge distillation in continual learning.}
Another solution to catastrophic forgetting is to preserve previous knowledge by self-distillation~\cite{iCaRL, DER++, kd1, Co2L, protoAug, OCM}. iCaRL~\cite{iCaRL} constrains changes of learned knowledge by distillation and employs class prototypes for nearest neighbor prediction. Co$^2$L~\cite{Co2L} proposes a self-distillation loss to preserve learned features. PASS~\cite{protoAug} maintains the decision boundaries of old classes by distilling old prototypes.
However, it is hard to distill useful knowledge when previous models are not learned well.
In contrast, we propose a general feedback mechanism to enhance the discrimination of classes that are prone to misclassification, which overcomes the limitations on knowledge distillation.

\paragraph{Prototypes in continual learning.}
Some previous methods~\cite{iCaRL, SCR, protoAug} attempt to utilize prototypes to mitigate catastrophic forgetting. As mentioned above, iCaRL and SCR employ class prototypes as classifiers, and PASS distills old prototypes to retain learned knowledge. Nevertheless, computing prototypes with all samples is extremely expensive for training. There are also some works considering the use of prototypes in the online scenario. CoPE~\cite{online_pro_ema} designs the prototypes with a high momentum-based update for each observed batch. A recent work~\cite{online_pro_accum} estimates class prototypes on all seen data using mean update criteria. However, regardless of momentum update or mean update, accumulating previous features as prototypes may be detrimental to future learning, since the features learned in old classes may not be discriminative when encountering new classes due to shortcut learning. 
In contrast, the proposed online prototypes only utilize the data visible at the current time step, which significantly decreases the computational cost and 
is more suitable for online CL.

\paragraph{Contrastive learning.}
Inspired by breakthroughs in self-supervised learning~\cite{CPC, MoCo, SimCLR, BYOL, SwAV, ProPos}, many studies~\cite{SCR, ER_AML, OCM, Co2L, online_pro_accum} in CL use contrastive learning to learn generalized features. An early work~\cite{ssl4onlineCL} analyzes and reveals the impact of contrastive learning on online CL. Among them, the work most related to ours is PCL~\cite{PCL}, which calculates infoNCE loss~\cite{CPC} between instance and prototype.
The most significant difference is that the loss in \methodname only considers online prototypes, and there is no involvement of instances. 
Please refer to Appendix~\ref{appendix:PCL} for detailed comparisons between our \methodname and PCL.

\section{Method}
Fig.~\ref{fig:framework} presents the illustration of the proposed \frameworkName.
In this section,  
we start by providing the problem definition of online CIL. Then, we describe the definition of the online prototype, the proposed online prototype equilibrium, and the proposed adaptive prototypical feedback. Finally, we propose an online prototype learning framework.

\subsection{Problem Definition}
Formally, online CIL considers a continuous sequence of tasks from a single-pass data stream $\mathfrak{D}=\left\{\mathcal{D}_1, \ldots, \mathcal{D}_T \right\} $, where $\mathcal{D}_t = \left\{ x_{i}, y_{i} \right\} ^{N_t}_{i=1} $ is the dataset of task $t$, and $T$ is the total number of tasks. Dataset $\mathcal{D}_t$ contains $N_t$ labeled samples, $y_{i}$ is the class label of sample $x_{i}$ and $y_{i} \in \mathcal{C}_t$, where $\mathcal{C}_t$ is the class set of task $t$ and the class sets of different tasks are disjoint. 
For replay-based methods, a memory bank is used to store a small subset of seen data, and we also maintain a memory bank $\mathcal{M}$ in our method.
At each time step of task $t$, the model receives a mini-batch data $X \cup X^\mathrm{b}$ for training, where $X$ and $X^\mathrm{b}$ are drawn from the i.i.d distribution $\mathcal{D}_t$ and the memory bank $\mathcal{M}$, respectively. 
Moreover, we adopt the single-head evaluation setup~\cite{EWC}, where a unified classifier must choose labels from all seen classes at inference due to unavailable task identifiers. 
The goal of online CIL is to train a unified model on data seen only once while predicting well on both new and old classes.

\subsection{Online Prototype Definition}
Prior to introducing the online prototypes, we first present the network architecture of our \frameworkName. Suppose that the model consists of three components: an encoder network $f$, a projection head $g$, and a classifier $\varphi$. Each sample $x$ in incoming data $X$ (a mini-batch data from new classes) is mapped to a projected vectorial embedding (representation) $\mathbf{z}$ by encoder $f$ and projector $g$:
\begin{align}
\label{eq:cal_z}
    \mathbf{z} = g(f(\operatorname{aug}(x);\theta_f);\theta_g),
\end{align}
where $\operatorname{aug}$ represents the data augmentation operation, $\theta_f$ and $\theta_g$ represent the parameters of $f$ and $g$, respectively, and $\mathbf{z}$ is $\ell_2$-normalized. 
Similar to Eq.~\eqref{eq:cal_z}, we use $\mathbf{z}^\mathrm{b}$ to denote the representation of replay data $X^\mathrm{b}$ (a mini-batch data from seen classes in the memory bank). 

At each time step of task $t$, the online prototype of each class is defined as the mean representation in a mini-batch:
\begin{align}
\label{eq:cal_p}
    \mathbf{p}_i = \frac{1}{n_i}\sum\nolimits_j\mathbf{z}_j\cdot \mathbbm{1}\{y_j = i\},
\end{align}
where $n_i$ is the number of samples for class $i$ in a mini-batch, and $\mathbbm{1}$ is the indicator function. 
We can get a set of $K$ online prototypes  in $X$, $\mathcal{P} = \left\{ \mathbf{p}_{i} \right\} ^{K}_{i=1}$, and a set of $K^\mathrm{b}$ online prototypes in $X^\mathrm{b}$, $\mathcal{P}^\mathrm{b} = \left\{ \mathbf{p}_i^\mathrm{b} \right\} ^{K^\mathrm{b}}_{i=1}$.
Note that $K = |\mathcal{P}| \leq |\mathcal{C}_t|$ and $K^\mathrm{b} = |\mathcal{P}^\mathrm{b}| \leq \sum_{i=1}^{t}|\mathcal{C}_i| $, where $|\cdot|$ denotes the cardinal number.

\subsection{Online Prototype Equilibrium}
The introduced online prototypes can provide representative features and avoid class-unrelated information.  
These characteristics are exactly the key to counteracting shortcut learning in online CL.
Besides, maintaining the discrimination among seen classes is also essential to mitigate catastrophic forgetting.
Based on these, we attempt to learn representative features of each class by pulling online prototypes $\mathcal{P}$ and their augmented views $\widehat{\mathcal{P}}$ closer in the embedding space, and learn discriminative features between classes by pushing online prototypes of different classes away, formally defined as a contrastive loss:
\begin{align}
\label{eq:proto_infoNCE}
    \ell(\mathcal{P},\widehat{\mathcal{P}})\!=\!
    \frac{-1}{|\mathcal{P}|}\sum_{i=1}^{|\mathcal{P}|}\!\log\! 
    \tfrac
    {\exp \big(\tfrac{{\mathbf{p}_i^\mathrm{T} \widehat{\mathbf{p}}_i}}{\tau}\big)}
    {
    \sum\limits_{j} \exp \big(\tfrac{{\mathbf{p}_i^\mathrm{T} \widehat{\mathbf{p}}_j}}{\tau}\big)
    +\!
    \sum\limits_{\substack{j \neq i}} \exp \big(\tfrac{{\mathbf{p}_i^\mathrm{T} \mathbf{p}_j}}{\tau}\big) 
    },
\end{align}
where 
$\tau$ is the temperature hyper-parameter, 
$\mathcal{P}$ and $\widehat{\mathcal{P}}$ are $\ell_2$-normalized. To compute the contrastive loss across all positive pairs in both $(\mathcal{P}, \widehat{\mathcal{P}})$ and $(\widehat{\mathcal{P}}, \mathcal{P})$, we define $\mathcal{L}_{\mathrm{pro}}$ as the final contrastive loss over online prototypes:
\begin{align}
    \mathcal{L}_{\mathrm{pro}}(\mathcal{P},\widehat{\mathcal{P}}) = 
    \frac{1}{2}
    \left[\ell(\mathcal{P}, \widehat{\mathcal{P}}) + \ell(\widehat{\mathcal{P}}, \mathcal{P})\right].
\end{align}

Considering the learning of new classes and the consolidation of learned knowledge simultaneously in online CL, we propose Online Prototype Equilibrium (\methodname) to 
learn representative and discriminative features on both new and seen classes by employing $\mathcal{L}_{\mathrm{pro}}$:
\begin{equation}
    \begin{aligned}
    \mathcal{L}_{\mathrm{\methodname}}
    &=
    \mathcal{L}^{\mathrm{new}}_{\mathrm{pro}}(\mathcal{P},\widehat{\mathcal{P}}) + \mathcal{L}^{\mathrm{seen}}_{\mathrm{pro}}(\mathcal{P}^\mathrm{b},\widehat{\mathcal{P}}^\mathrm{b}),
    \end{aligned}
\end{equation}
where
$\mathcal{L}^{\mathrm{new}}_{\mathrm{pro}}$ focuses on learning knowledge from \emph{new} classes, and $\mathcal{L}^{\mathrm{seen}}_{\mathrm{pro}}$ is dedicated to preserving learned knowledge of all \emph{seen} classes.
\textit{This process is similar to a zero-sum game, 
and \methodname aims to achieve the equilibrium to play a win-win game.}
Concretely,
as the model learns, the knowledge of new classes is gained and added to the prototypes over the memory bank $\mathcal{M}$, causing $\mathcal{L}^{\mathrm{seen}}_{\mathrm{pro}}$ gradually changes to the equilibrium that separates all seen classes well, including new ones. 
This variation is crucial to mitigate forgetting and is consistent with the goal of CIL.

\subsection{Adaptive Prototypical Feedback} 
Although \methodname can bring an overall equilibrium, it tends to treat each class \emph{equally}. 
In fact, the degree of confusion varies among classes, 
and the model should focus purposefully on confused classes to consolidate learned knowledge. 
To this end, we propose Adaptive Prototypical Feedback (\dataaugname) with the feedback of online prototypes to sense the classes that are prone to be misclassified and then enhance their decision boundaries.
 
For each class pair in the memory bank $\mathcal{M}$, \dataaugname calculates the distances between online prototypes of all seen classes from the previous time step, showing the class confusion status by these distances. The closer the two prototypes are, the easier to be misclassified. 
Based on this analysis, 
our idea is to enhance the boundaries for those classes. Therefore, we convert the prototype distance matrix to a probability distribution $P$ over the classes via a symmetric Gaussian kernel, defined as follows:
\begin{align}
\label{eq:cal_d}
    P_{i, j} \propto \exp (-\| \mathbf{p}_i^\mathrm{b} - \mathbf{p}_j^\mathrm{b} \|_2^2),
\end{align}
where $i,j \in \{ 1, \ldots, |\mathcal{P}^\mathrm{b}| \}$ and $i \neq j$. 
Then, 
all probabilities are normalized to a probability mass function that sums to one.
\dataaugname returns probabilities to $\mathcal{M}$ for guiding the next sampling process and enhancing decision boundaries of easily misclassified classes.

Our adaptive prototypical feedback is implemented as a sampling-based mixup. Specifically, 
\dataaugname adaptively selects more samples from easily misclassified classes in $\mathcal{M}$ for mixup~\cite{Mixup} according to the probability distribution $P$. 
Considering not over-penalizing the equilibrium of current online prototypes, we introduce a two-stage sampling strategy for replay data $X^\mathrm{b}$ of size $m$. 
First, we select $n_{\mathrm{\dataaugname}}$ samples  
with $P$, and a larger $P_{a,b}$ means more sampling from classes $a$ and $b$. Here, $n_{\mathrm{\dataaugname}} = \alpha \cdot m$, and $\alpha$ is the ratio of \dataaugname.
Second, the remaining $m-n_{\mathrm{\dataaugname}}$ samples are uniformly randomly selected from the entire memory bank to avoid the model only focusing on easily misclassified classes and disrupting the established equilibrium.

\subsection{Overall Framework of \frameworkName}
The overall structure of \frameworkName is shown in Fig.~\ref{fig:framework}. \frameworkName comprises two key components based on proposed online prototypes: Online Prototype Equilibrium (\methodname) and Adaptive Prototypical Feedback (\dataaugname). 
With the two components, 
the model can learn representative features against shortcut learning, and 
all seen classes maintain discriminative when learning new classes. 
However, classes may not be compact, because the online prototypes cannot cover full instance-level information.
To further achieve intra-class compactness, 
we employ supervised contrastive learning~\cite{SupCL} to learn instance-wise representations:
\begin{equation}
\begin{aligned}
    \mathcal{L}_{\mathrm{INS}}
    &=
    \sum_{i=1}^{2N} \frac{-1}{\left|I_i\right|} \sum_{j \in I_i} \log \frac{\exp \left(\mathrm{sim}(\mathbf{z}_i, \mathbf{z}_j) / \tau^{\prime}\right)}{\sum\limits_{k \neq i} \exp \left(\mathrm{sim}(\mathbf{z}_i, \mathbf{z}_k) / \tau^{\prime}\right)}
    \\
    &+
    \sum_{i=1}^{2m} \frac{-1}{\left|I_i^{\mathrm{b}}\right|} \sum_{j \in I_i^{\mathrm{b}}} \log \frac{\exp (\mathrm{sim}(\mathbf{z}_i^{\mathrm{b}}, \mathbf{z}_j^{\mathrm{b}}) / \tau^{\prime})}{\sum\limits_{k \neq i} \exp \left(\mathrm{sim}(\mathbf{z}_i^{\mathrm{b}}, \mathbf{z}_k^{\mathrm{b}}) / \tau^{\prime}\right)},
\end{aligned}
\end{equation}
where $I_i=\left\{j \in\{1, \ldots, 2 N\} \mid j \neq i, y_j=y_i\right\}$ and $I_i^\mathrm{{b}}=\left\{j \in\{1, \ldots, 2m\} \mid j \neq i, y_j^\mathrm{b}=y_i^\mathrm{b}\right\}$ are the set of positive samples indexes to $\mathbf{z}_i$ and $\mathbf{z}_i^\mathrm{{b}}$, respectively. $y_i^\mathrm{b}$ is the class label of input $x_i^\mathrm{b}$ from $X^\mathrm{b}$. $N$ is the batch size of $X$. $\tau^{\prime}$ is the temperature hyperparameter.
The similarity function $\mathrm{sim}$ is computed in the same way as Eq.~(9) in OCM~\cite{OCM}.

Thus, the total loss of our \frameworkName framework is given as:
\begin{align}
    \mathcal{L}_{\mathrm{\frameworkName}}=\mathcal{L}_{\mathrm{\methodname}} + \mathcal{L}_{\mathrm{INS}} + \mathcal{L}_{\mathrm{CE}},
\end{align}
where $\mathcal{L}_{\mathrm{CE}} = \mathrm{CE}(y^\mathrm{b}, \varphi(f(\operatorname{aug}(x^\mathrm{b}))))$ is the cross-entropy loss; see Appendix~\ref{appendix:algorithm} for detailed training algorithms.

Following other replay-based methods~\cite{ER, SCR, OCM}, we update the memory bank in each time step by uniformly randomly selecting samples from $X$ to push into $\mathcal{M}$ and, if $\mathcal{M}$ is full, pulling an equal number of samples out of $\mathcal{M}$.

\begin{figure*}
    \centering
    \includegraphics[width=1.0\linewidth]{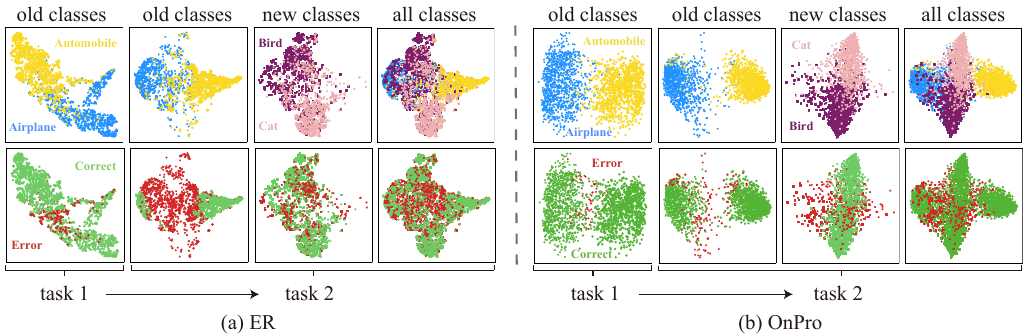}
    \caption{$t$-SNE~\cite{tsne} visualizations of features learned from ER and \frameworkName on the test set of CIFAR-10.
    When learning new classes, ER suffers serious class confusion probably because shortcut learning. In contrast, \frameworkName significantly mitigates the forgetting.
    }
    \label{fig:tsne_motivation}
\end{figure*}
\begin{table*}[ht]
\small
\begin{center}
\resizebox{\linewidth}{!}{
\begin{tabular}{rrrrrrrrrrrr}
\shline
\multirow{2}{*}{Method}  & \multicolumn{3}{c}{CIFAR-10}   && \multicolumn{3}{c}{CIFAR-100}  && \multicolumn{3}{c}{TinyImageNet} \\ \cline{2-4}\cline{6-8}\cline{10-12}
       & $M=0.1k$   & $M=0.2k$   & $M=0.5k$     && $M=0.5k$     & $M=1k$     & $M=2k$     && $M=1k$      & $M=2k$ & $M=4k$   \\ \midrule
iCaRL~\cite{iCaRL}    & 31.0\std{$\pm$1.2} & 33.9\std{$\pm$0.9} & 42.0\std{$\pm$0.9} && 12.8\std{$\pm$0.4}  & 16.5\std{$\pm$0.4}  & 17.6\std{$\pm$0.5} && 5.0\std{$\pm$0.3}   & 6.6\std{$\pm$0.4} & 7.8\std{$\pm$0.4} \\ 
DER++~\cite{DER++}   & 31.5\std{$\pm$2.9} & 39.7\std{$\pm$2.7} & 50.9\std{$\pm$1.8} && 16.0\std{$\pm$0.6}  & 21.4\std{$\pm$0.9}  & 23.9\std{$\pm$1.0} && 3.7\std{$\pm$0.4} & 5.1\std{$\pm$0.8} & 6.8\std{$\pm$0.6} \\ 
PASS~\cite{protoAug}    & 33.7\std{$\pm$2.2} & 33.7\std{$\pm$2.2} & 33.7\std{$\pm$2.2} && 7.5\std{$\pm$0.7}  & 7.5\std{$\pm$0.7}  & 7.5\std{$\pm$0.7} && 0.5\std{$\pm$0.1}   & 0.5\std{$\pm$0.1} & 0.5\std{$\pm$0.1} \\ 
\hline
AGEM~\cite{AGEM}    & 17.7\std{$\pm$0.3} & 17.5\std{$\pm$0.3} & 17.5\std{$\pm$0.2} && 5.8\std{$\pm$0.1}  & 5.9\std{$\pm$0.1}  & 5.8\std{$\pm$0.1} && 0.8\std{$\pm$0.1}   & 0.8\std{$\pm$0.1} & 0.8\std{$\pm$0.1} \\ 
GSS~\cite{GSS}     & 18.4\std{$\pm$0.2} & 19.4\std{$\pm$0.7} & 25.2\std{$\pm$0.9} && 8.1\std{$\pm$0.2}  & 9.4\std{$\pm$0.5}  & 10.1\std{$\pm$0.8} && 1.1\std{$\pm$0.1}   & 1.5\std{$\pm$0.1} & 2.4\std{$\pm$0.4} \\ 
ER~\cite{ER}      & 19.4\std{$\pm$0.6} & 20.9\std{$\pm$0.9} & 26.0\std{$\pm$1.2} && 8.7\std{$\pm$0.3}  & 9.9\std{$\pm$0.5}  & 10.7\std{$\pm$0.8} && 1.2\std{$\pm$0.1}   & 1.5\std{$\pm$0.2} & 2.0\std{$\pm$0.2} \\ 
MIR~\cite{MIR}     & 20.7\std{$\pm$0.7} & 23.5\std{$\pm$0.8} & 29.9\std{$\pm$1.2} && 9.7\std{$\pm$0.3}  & 11.2\std{$\pm$0.4}  & 13.0\std{$\pm$0.7} && 1.4\std{$\pm$0.1}   & 1.9\std{$\pm$0.2} & 2.9\std{$\pm$0.3} \\ 
GDumb~\cite{GDumb}   & 23.3\std{$\pm$1.3} & 27.1\std{$\pm$0.7} & 34.0\std{$\pm$0.8} && 8.2\std{$\pm$0.2}  & 11.0\std{$\pm$0.4}  & 15.3\std{$\pm$0.3} && 4.6\std{$\pm$0.3}   & 6.6\std{$\pm$0.2} & 10.0\std{$\pm$0.3} \\ 
ASER~\cite{ASER}   & 20.0\std{$\pm$1.0} & 22.8\std{$\pm$0.6} & 31.6\std{$\pm$1.1} && 11.0\std{$\pm$0.3}  & 13.5\std{$\pm$0.3}  & 17.6\std{$\pm$0.4} && 2.2\std{$\pm$0.1}   & 4.2\std{$\pm$0.6} & 8.4\std{$\pm$0.7} \\ 
SCR~\cite{SCR}     & 40.2\std{$\pm$1.3} & 48.5\std{$\pm$1.5} & 59.1\std{$\pm$1.3} && 19.3\std{$\pm$0.6}  & 26.5\std{$\pm$0.5}  & 32.7\std{$\pm$0.3} && 8.9\std{$\pm$0.3}   & 14.7\std{$\pm$0.3} & 19.5\std{$\pm$0.3} \\ 
CoPE~\cite{online_pro_ema}  & 33.5\std{$\pm$3.2} & 37.3\std{$\pm$2.2} & 42.9\std{$\pm$3.5} && 11.6\std{$\pm$0.7}  & 14.6\std{$\pm$1.3}  & 16.8\std{$\pm$0.9} && 2.1\std{$\pm$0.3}   & 2.3\std{$\pm$0.4} & 2.5\std{$\pm$0.3} \\
DVC~\cite{DVC} & 35.2\std{$\pm$1.7}  & 41.6\std{$\pm$2.7} & 53.8\std{$\pm$2.2} &&  15.4\std{$\pm$0.7} & 20.3\std{$\pm$1.0} & 25.2\std{$\pm$1.6} && 4.9\std{$\pm$0.6} &  7.5\std{$\pm$0.5} & 10.9\std{$\pm$1.1} \\ 
OCM~\cite{OCM} & 47.5\std{$\pm$1.7}  & 59.6\std{$\pm$0.4} & 70.1\std{$\pm$1.5} && 19.7\std{$\pm$0.5} & 27.4\std{$\pm$0.3} & 34.4\std{$\pm$0.5} && 10.8\std{$\pm$0.4} & 15.4\std{$\pm$0.4} & 20.9\std{$\pm$0.7} \\ 
\hline
\frameworkName (\textbf{ours}) & \textbf{57.8}\std{$\pm$1.1} & \textbf{65.5}\std{$\pm$1.0} & \textbf{72.6}\std{$\pm$0.8} && \textbf{22.7}\std{$\pm$0.7} & \textbf{30.0}\std{$\pm$0.4} & \textbf{35.9}\std{$\pm$0.6} && \textbf{11.9}\std{$\pm$0.3} & \textbf{16.9}\std{$\pm$0.4} &  \textbf{22.1}\std{$\pm$0.4}
\\ 
\shline
\end{tabular}
}
\end{center}
\caption{Average Accuracy~(higher is better) on three benckmark datasets with different memory bank sizes $M$. All results are the average and standard deviation of 15 runs.}
\label{tab:acc}
\end{table*}

\section{Experiments}
\subsection{Experimental Setup}
\paragraph{Datasets.}
We use three image classification benchmark datasets, including \textbf{CIFAR-10}~\cite{cifar10_100}, \textbf{CIFAR-100}~\cite{cifar10_100}, and \textbf{TinyImageNet}~\cite{tinyImageNet}, to evaluate the performance of online CIL methods. 
Following~\cite{ASER, SCR, DVC}, we split CIFAR-10 into 5 disjoint tasks, where each task has 2 disjoint classes, 10,000 samples for training, and 2,000 samples for testing, and split CIFAR-100 into 10 disjoint tasks, where each task has 10 disjoint classes, 5,000 samples for training, and 1,000 samples for testing.
Following~\cite{OCM}, we split TinyImageNet into 100 disjoint tasks, where each task has 2 disjoint classes, 1,000 samples for training, and 100 samples for testing.
Note that the order of tasks is fixed in all experimental settings.

\paragraph{Baselines.}
We compare our \frameworkName with 13 baselines, including 10 replay-based online CL baselines: {AGEM}~\cite{AGEM}, {MIR}~\cite{MIR}, {GSS}~\cite{GSS}, {ER}~\cite{ER}, {GDumb}~\cite{GDumb}, {ASER}~\cite{ASER}, {SCR}~\cite{SCR}, {CoPE}~\cite{online_pro_ema}, {DVC}~\cite{DVC}, and {OCM}~\cite{OCM}; 3 offline CL baselines that use knowledge distillation by running them in one epoch: {iCaRL}~\cite{iCaRL}, {DER++}~\cite{DER++}, and PASS~\cite{protoAug}. Note that PASS is a non-exemplar method.

\paragraph{Evaluation metrics.}
We use Average Accuracy and Average Forgetting~\cite{ASER, DVC} to measure the performance of our framework in online CIL. Average Accuracy evaluates the accuracy of the test sets from all seen tasks, defined as $\text {Average Accuracy} =\frac{1}{T} \sum_{j=1}^T a_{T, j},$
where $a_{i, j}$ is the accuracy on task $j$ after the model is trained from task $1$ to $i$.
Average Forgetting represents how much the model forgets about each task after being trained on the final task, defined as
$\text { Average Forgetting } =\frac{1}{T-1} \sum_{j=1}^{T-1} f_{T, j}, 
\text { where } f_{i, j}=\max _{k \in\{1, \ldots, i-1\}} a_{k, j}-a_{i, j}.$

\paragraph{Implementation details.}
We use ResNet18~\cite{ResNet} as the backbone $f$ and a linear layer as the projection head $g$ like~\cite{SCR, OCM, Co2L}; the hidden dim in $g$ is set to 128 as~\cite{SimCLR}. We also employ a linear layer as the classifier $\varphi$. We train the model from scratch with Adam optimizer and an initial learning rate of $5\times10^{-4}$ for all datasets. The weight decay is set to $1.0\times10^{-4}$. Following~\cite{ASER, DVC}, we set the batch size $N$ as 10, and following~\cite{OCM} the replay batch size $m$ is set to 64. 
For CIFAR-10, we set the ratio of \dataaugname $\alpha = 0.25$. For CIFAR-100 and TinyImageNet, $\alpha$ is set to $0.1$. The temperature $\tau = 0.5$ and $\tau^{\prime} = 0.07$.
For baselines, we also use ResNet18 as their backbone and set the same batch size and replay batch size for fair comparisons.
We reproduce all baselines in the same environment with their source code and default settings; see Appendix~\ref{appendix:baselines} for implementation details about all baselines.
We report the average results across 15 runs for all experiments.

\paragraph{Data augmentation.}
Similar to data augmentations used in SimCLR~\cite{SimCLR}, we use resized-crop, horizontal-flip, and gray-scale as our data augmentations. For all baselines, we also use these augmentations. In addition, for DER++\cite{DER++}, SCR~\cite{SCR}, and DVC~\cite{DVC}, we follow their default settings and use their own extra data augmentations. OCM~\cite{OCM} uses extra rotation augmentations, which are also used in \frameworkName.

\subsection{Motivation Justification}
\label{pre_exp}
\paragraph{Shortcut learning in online CL.}
Shortcut learning is severe in online CL since the model cannot learn sufficient representative features due to the single-pass data stream. To intuitively demonstrate this issue,  
we conduct GradCAM++~\cite{Grad-cam++} on the training set of CIFAR-10 ($M=0.2k$) after the model is trained incrementally, as shown in Fig.~\ref{fig:heatmap}.
Each row in Fig.~\ref{fig:heatmap} represents a task with two classes.
We can observe that although ER and DVC predict the correct class, the models actually take shortcuts and focus on some object-unrelated features. 
An interesting phenomenon is that ER tends to take shortcuts in each task. For example, ER learns the sky on both the airplane class in task 1 (the first row) and the bird class in task 2 (the second row) . Thus, ER forgets almost all the knowledge of the old classes.  
DVC maximizes the mutual information between instances like contrastive learning~\cite{SimCLR, MoCo}, which only partially alleviates shortcut learning in online CL. 
In contrast, \frameworkName focuses on the representative features of the objects themselves. The results confirm that learning representative features is crucial against shortcut learning; see Appendix~\ref{appendix:more_visual} for more visual explanations.

\begin{table*}[htbp]
\small
\begin{center}
\resizebox{\linewidth}{!}{
\begin{tabular}{rrrrrrrrrrrr}
\shline
\multirow{2}{*}{Method}  & \multicolumn{3}{c}{CIFAR-10}   && \multicolumn{3}{c}{CIFAR-100}  && \multicolumn{3}{c}{TinyImageNet} \\ \cline{2-4}\cline{6-8}\cline{10-12}
       &  $M=0.1k$   &  $M=0.2k$   &  $M=0.5k$     &&  $M=0.5k$     &  $M=1k$     &  $M=2k$    &&  $M=1k$      &  $M=2k$ &  $M=4k$    \\ \midrule
iCaRL~\cite{iCaRL}    & 52.7\std{$\pm$1.0} & 49.3\std{$\pm$0.8} & 38.3\std{$\pm$0.9} && 16.5\std{$\pm$1.0}  & 11.2\std{$\pm$0.4}  & 10.4\std{$\pm$0.4} && 9.9\std{$\pm$0.5}   & 10.1\std{$\pm$0.5} & 9.7\std{$\pm$0.6} \\ 
DER++~\cite{DER++}   & 57.8\std{$\pm$4.1} & 46.7\std{$\pm$3.6} & 33.6\std{$\pm$3.5} && 41.0\std{$\pm$1.1} & 34.8\std{$\pm$1.1} & 33.2\std{$\pm$1.2} && 77.8\std{$\pm$1.0} & 74.9\std{$\pm$0.6} & 73.2\std{$\pm$0.8}  \\ 
PASS~\cite{protoAug}    & 21.2\std{$\pm$2.2} & 21.2\std{$\pm$2.2} & 21.2\std{$\pm$2.2} && 10.6\std{$\pm$0.9}  & 10.6\std{$\pm$0.9}  & 10.6\std{$\pm$0.9} && 27.0\std{$\pm$2.4}   & 27.0\std{$\pm$2.4} & 27.0\std{$\pm$2.4} \\ 
\hline
AGEM~\cite{AGEM}    & 64.8\std{$\pm$0.7} & 64.8\std{$\pm$0.7} & 64.5\std{$\pm$0.5} && 41.7\std{$\pm$0.8} & 41.8\std{$\pm$0.7} & 41.7\std{$\pm$0.6} && 73.9\std{$\pm$0.7} & 73.1\std{$\pm$0.7} & 72.9\std{$\pm$0.5} \\ 
GSS~\cite{GSS}     & 67.1\std{$\pm$0.6} & 65.8\std{$\pm$0.6} & 61.2\std{$\pm$1.2} && 48.7\std{$\pm$0.8} & 46.7\std{$\pm$1.3} & 44.7\std{$\pm$1.1} && 78.9\std{$\pm$0.7} & 77.0\std{$\pm$0.5} & 75.2\std{$\pm$0.7} \\ 
ER~\cite{ER}      & 64.7\std{$\pm$1.1} & 62.9\std{$\pm$1.0} & 57.5\std{$\pm$1.8} && 47.0\std{$\pm$1.0} & 46.4\std{$\pm$0.8} & 44.7\std{$\pm$1.5} && 79.1\std{$\pm$0.6} & 77.7\std{$\pm$0.6} & 76.3\std{$\pm$0.5} \\ 
MIR~\cite{MIR}     & 62.6\std{$\pm$1.0} & 58.5\std{$\pm$1.4} & 51.1\std{$\pm$1.1} && 45.7\std{$\pm$0.9} & 44.2\std{$\pm$1.3} & 42.3\std{$\pm$1.0} && 75.3\std{$\pm$0.9} & 71.5\std{$\pm$1.0} & 66.8\std{$\pm$0.8} \\ 
GDumb~\cite{GDumb}   & 28.5\std{$\pm$1.4} & 28.4\std{$\pm$1.0} & 28.1\std{$\pm$1.0} && 25.0\std{$\pm$0.4} & 23.2\std{$\pm$0.4} & 20.7\std{$\pm$0.3}  && 22.7\std{$\pm$0.3} & 18.4\std{$\pm$0.2} & 17.0\std{$\pm$0.2} \\
ASER~\cite{ASER}    & 64.8\std{$\pm$1.0} & 62.6\std{$\pm$1.1} & 53.2\std{$\pm$1.5} && 52.8\std{$\pm$0.8} & 50.4\std{$\pm$0.9} & 46.8\std{$\pm$0.7} && 78.9\std{$\pm$0.5} & 75.4\std{$\pm$0.7} & 68.2\std{$\pm$1.1} \\ 
SCR~\cite{SCR}     & 43.2\std{$\pm$1.5} & 35.5\std{$\pm$1.8} & 24.1\std{$\pm$1.0} && 29.3\std{$\pm$0.9} & 20.4\std{$\pm$0.6} & 11.5\std{$\pm$0.6} && 44.8\std{$\pm$0.6} & 26.8\std{$\pm$0.5} & 20.1\std{$\pm$0.4} \\ 
CoPE~\cite{online_pro_ema}  & 49.7\std{$\pm$1.6} & 45.7\std{$\pm$1.5} & 39.4\std{$\pm$1.8} && 25.6\std{$\pm$0.9}  & 17.8\std{$\pm$1.3}  & 14.4\std{$\pm$0.8} && 11.9\std{$\pm$0.6}   & 10.9\std{$\pm$0.4} & 9.7\std{$\pm$0.4} \\
DVC~\cite{DVC} & 40.2\std{$\pm$2.6} & 31.4\std{$\pm$4.1} & 21.2\std{$\pm$2.8} && 32.0\std{$\pm$0.9} & 32.7\std{$\pm$2.0} & 28.0\std{$\pm$2.2} && 59.8\std{$\pm$2.2} & 52.9\std{$\pm$1.3} & 45.1\std{$\pm$1.9} \\
OCM~\cite{OCM} & 35.5\std{$\pm$2.4} & 23.9\std{$\pm$1.4} & 13.5\std{$\pm$1.5} && 18.3\std{$\pm$0.9} & 15.2\std{$\pm$1.0} & 10.8\std{$\pm$0.6} && 23.6\std{$\pm$0.5} & 26.2\std{$\pm$0.5}  & 23.8\std{$\pm$1.0} \\ 
\hline
{\frameworkName} (\textbf{ours})   & 23.2\std{$\pm$1.3} & 17.6\std{$\pm$1.4} & 12.5\std{$\pm$0.7} && 
15.0\std{$\pm$0.8} & 10.4\std{$\pm$0.5} & 6.1\std{$\pm$0.6} && 21.3\std{$\pm$0.5} & 17.4\std{$\pm$0.4} & 16.8\std{$\pm$0.4} \\
\shline
\end{tabular}
}
\end{center}
\caption{Average Forgetting~(lower is better) on three benckmark datasets. All results are the average and standard deviation of 15 runs.}
\label{tab:forget}
\end{table*}

\begin{figure*}[htp]
  \centering
  \subfloat[Average incremental performance]{
    \includegraphics[width=0.55\linewidth]{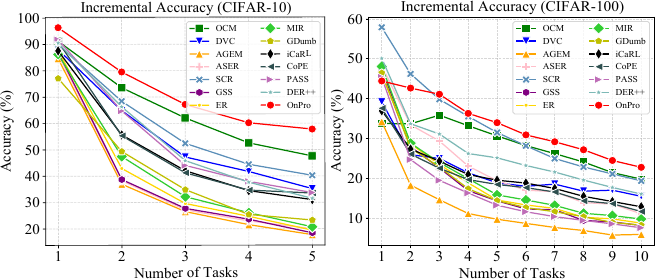}
    \label{fig:incrementalAcc}
  }
  \subfloat[Confusion matrix of OCM and \frameworkName]{
    \includegraphics[width=0.42\linewidth]{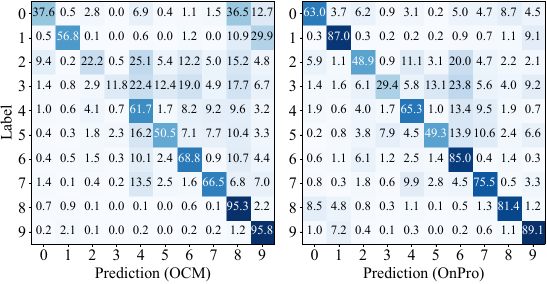}
    \label{fig:confusionMatrix}
  }
  \caption{Incremental accuracy on tasks observed so far and confusion matrix of accuracy (\%) in the {test set} of CIFAR-10.}
  \label{fig:incrementalAcc_confusionMatrix}
\end{figure*}

\paragraph{Class confusion in online CL.}
Fig.~\ref{fig:tsne_motivation} provides the $t$-SNE~\cite{tsne} visualization results for ER and \frameworkName on the test set of CIFAR-10 ($M=0.2k$). 
We can draw intuitive observations as follows. 
(1) There is serious class confusion in ER.
When the new task (task 2) arrives, features learned in task 1 are not discriminative for task 2, leading to class confusion and decreased performance in old classes.
(2) Shortcut learning may cause class confusion. For example, the performance of ER decreases more on airplanes compared to automobiles, probably because birds in the new task have more similar backgrounds to airplanes, as shown in Fig.~\ref{fig:heatmap}.
(3) \frameworkName achieves better discrimination both on task 1 and task 2. The results demonstrate that \frameworkName can maintain discrimination of all seen classes and significantly mitigate forgetting by 
combining the proposed \methodname and \dataaugname.

\subsection{Results and Analysis}
\label{result}
\paragraph{Performance of average accuracy.}
Table~\ref{tab:acc} presents the results of average accuracy with different memory bank sizes ($M$) on three benchmark datasets. Our \frameworkName consistently outperforms all baselines on three datasets.
Remarkably, the performance improvement of \frameworkName is more significant when the memory bank size is relatively small; this is critical for online CL with limited resources. For example, compared to the second-best method OCM, \frameworkName achieves about 10$\%$ and 6$\%$ improvement on CIFAR-10 when $M$ is 100 and 200, respectively. 
The results show that our \frameworkName can learn more representative and discriminative features with a limited memory bank.
Compared to baselines that use knowledge distillation (iCaRL, DER++, PASS, OCM), our \frameworkName achieves better performance by leveraging the feedback of online prototypes.  
Besides, \frameworkName significantly outperforms PASS and CoPE that also use prototypes, showing that online prototypes are more suitable for online CL.

We find that the performance improvement tends to be gentle when $M$ increases.
The reason is that as $M$ increases, the samples in the memory bank become more diverse, and the model can extract sufficient information from massive samples to distinguish seen classes. 
In addition, many baselines perform poorly on CIFAR-100 and TinyImageNet due to a dramatic increase in the number of tasks. In contrast, \frameworkName still performs well and improves accuracy over the second best.

\paragraph{Performance of average forgetting.}
We report the Average Forgetting results of our \frameworkName and all baselines on three benchmark datasets in Table~\ref{tab:forget}. The results confirm that \frameworkName can effectively mitigate catastrophic forgetting. 
For CIFAR-10 and CIFAR-100, \frameworkName achieves the lowest average forgetting compared to all replay-based baselines. 
For TinyImageNet, our result is a little higher than iCaRL and CoPE but better than the latest methods DVC and OCM. 
The reason is that iCaRL uses a nearest class mean classifier, but we use softmax and FC layer during the test phase, and CoPE slowly updates prototypes with a high momentum.
However, as shown in Table~\ref{tab:acc}, \frameworkName provides more accurate classification results than iCaRL and CoPE. 
It is a fact that when the maximum accuracy of a task is small, the forgetting on this task is naturally rare, even if the model completely forgets what it learned.

\paragraph{Performance of each incremental step.}
We evaluate the average incremental performance~\cite{DER++, DVC} on CIFAR-10 ($M=0.1k$) and CIFAR-100 ($M=0.5k$), which indicates the accuracy over all seen tasks at each incremental step. 
Fig.~\ref{fig:incrementalAcc} shows that \frameworkName achieves better accuracy and effectively mitigates forgetting while the performance of most baselines degrades rapidly with the arrival of new classes.

\paragraph{Confusion matrices at the end of learning.}
We report the confusion matrices of our \frameworkName and the second-best method OCM, as shown in Fig.~\ref{fig:confusionMatrix}. 
After learning the last task (\ie, the last two classes), OCM forgets the knowledge of early tasks (classes 0 to 3). 
In contrast, \frameworkName performs relatively well in all classes, especially in the first task (classes 0 and 1), outperforming OCM by 27.8\% average improvements.
The results show that learning representative and discriminative features is crucial to mitigate catastrophic forgetting; see Appendix~\ref{appendix:extra_exp} for extra experimental results.

\subsection{Ablation Studies}
\label{ablation}

\begin{table}[t]
\small
\begin{center}
\begin{tabular}{ccccc}
\shline
\multirow{2}{*}{{Method}} & {CIFAR-10}&{CIFAR-100} \\
& Acc $\uparrow$(Forget $\downarrow$) & Acc $\uparrow$(Forget $\downarrow$) \\ 
\midrule
baseline & 46.4\std{$\pm$1.2}(36.0\std{$\pm$}2.1) & 18.8\std{$\pm$0.8}(18.5\std{$\pm$}0.7) \\
w/o \methodname & 53.1\std{$\pm$1.4}(24.7\std{$\pm$2.0}) & 19.3\std{$\pm$0.7}(15.9\std{$\pm$0.9}) \\
w/o \dataaugname & 52.0\std{$\pm$1.5}(34.6\std{$\pm$2.4}) & 21.5\std{$\pm$0.5}(16.3\std{$\pm$0.8}) \\ 
\hline
w/o $\mathcal{L}^{\mathrm{new}}_{\mathrm{pro}}$ & 54.8\std{$\pm$1.2}(\textbf{22.1}\std{$\pm$3.0}) & 19.6\std{$\pm$0.8}(19.9\std{$\pm$0.7}) \\
w/o $\mathcal{L}^{\mathrm{seen}}_{\mathrm{pro}}$ & 55.7\std{$\pm$1.4}(25.5\std{$\pm$1.5}) & 20.1\std{$\pm$0.4}(16.2\std{$\pm$0.6}) \\ 
$\mathcal{L}^{\mathrm{seen}}_{\mathrm{pro}}$ w/o $\mathcal{C}^\mathrm{new}$ & 56.2\std{$\pm$1.2}(26.4\std{$\pm$2.3}) & 20.8\std{$\pm$0.6}(17.9\std{$\pm$0.7}) \\ 
\hline
{\frameworkName} (\textbf{ours}) & \textbf{57.8}\std{$\pm$1.1}(23.2\std{$\pm$1.3}) & \textbf{22.7}\std{$\pm$0.7}(\textbf{15.0}\std{$\pm$0.8}) \\ 
\shline 
\end{tabular}
\end{center}
\caption{Ablation studies on CIFAR-10 ($M=0.1k$) and CIFAR-100 ($M=0.5k$). 
``baseline'' means $\mathcal{L}_\mathrm{INS}+\mathcal{L}_\mathrm{CE}$.
``$\mathcal{L}^{\mathrm{seen}}_{\mathrm{pro}}$ w/o $\mathcal{C}^\mathrm{new}$'' means $\mathcal{L}^{\mathrm{seen}}_{\mathrm{pro}}$ do not consider new classes in current task.
}
\label{tab:ablation}
\end{table}

\paragraph{Effects of each component.} Table~\ref{tab:ablation} presents the ablation results of each component. Obviously, \methodname and \dataaugname can consistently improve the average accuracy of classification. 
We can observe that the effect of \methodname is more significant on more tasks while \dataaugname plays a crucial role when the memory bank size is limited. Moreover, when combining \methodname and \dataaugname, the performance is further improved, which indicates that both can benefit from each other. For example, \dataaugname boosts \methodname by about 6$\%$ improvements on CIFAR-10 ($M=0.1k$), and the performance of \dataaugname is improved by about 3$\%$ on CIFAR-100 ($M=0.5k$) by combining \methodname.

\paragraph{Equilibrium in \methodname.}
When learning new classes, the data of new classes is involved in both $\mathcal{L}^{\mathrm{new}}_{\mathrm{pro}}$ and $\mathcal{L}^{\mathrm{seen}}_{\mathrm{pro}}$ of \methodname, where $\mathcal{L}^{\mathrm{new}}_{\mathrm{pro}}$ only focuses on learning new knowledge while $\mathcal{L}^{\mathrm{seen}}_{\mathrm{pro}}$ tends to alleviate forgetting on seen classes.
To explore the best way of learning new classes, we consider three scenarios for \methodname in Table~\ref{tab:ablation}.
The results show that only learning new knowledge (w/o $\mathcal{L}^{\mathrm{seen}}_{\mathrm{pro}}$) or only consolidating the previous knowledge (w/o $\mathcal{L}^{\mathrm{new}}_{\mathrm{pro}}$) can significantly degrade the performance, which indicates that both are indispensable for online CL.
Furthermore, when $\mathcal{L}^{\mathrm{seen}}_{\mathrm{pro}}$ only considers old classes and ignores new classes ($\mathcal{L}^{\mathrm{seen}}_{\mathrm{pro}}$ w/o $\mathcal{C}^\mathrm{new}$), the performance also decreases. These results show that the equilibrium of all seen classes (\methodname) can achieve the best performance and is crucial for online CL.

\paragraph{Effects of \dataaugname.} 
To verify the advantage of \dataaugname, we compare it with the completely random mixup
in Table~\ref{tab:ablation_mixup}.
\begin{table}
\small
\begin{center}
\begin{tabular}{c|rrr}
\shline
\multicolumn{1}{c|}{Method}       & ${M=0.1k}$   & ${M=0.2k}$   & ${M=0.5k}$     \\ \hline
Random & 53.5\std{$\pm$2.7} & 62.9\std{$\pm$2.5} & 70.8\std{$\pm$2.2} \\
\dataaugname (\textbf{ours})  & \textbf{57.8}\std{$\pm$1.1} & \textbf{65.5}\std{$\pm$1.0} & \textbf{72.6}\std{$\pm$0.8} \\ 
\shline
\end{tabular}
\end{center}
\caption{Comparison of Random Mixup and \dataaugname on CIFAR-10. 
}
\label{tab:ablation_mixup}
\end{table}
\dataaugname outperforms random mixup in all three scenarios. Notably, \dataaugname works significantly when the memory bank size is small, which shows that the feedback can prevent class confusion due to a restricted memory bank; see Appendix~\ref{appendix:ablations} for extra ablation studies.

\subsection{Validation of Online Prototypes}
\label{prove_onlinePrototypes}
\begin{figure}
    \centering
    \includegraphics[width=1.0\linewidth]{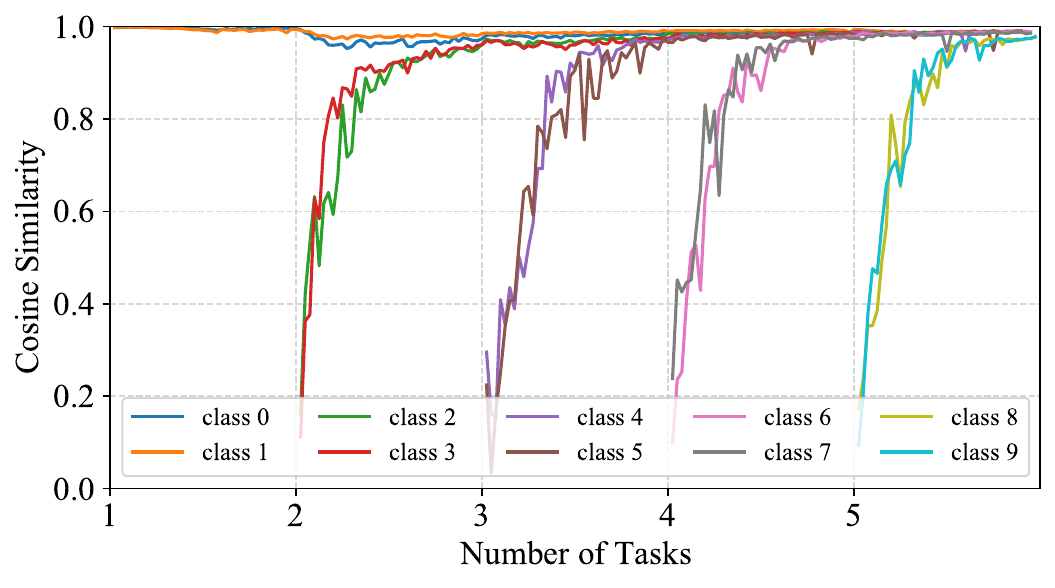}
    \caption{The cosine similarity between online prototypes and prototypes of the entire memory bank.}
    \label{fig:cosine_similarity}
\end{figure}
Fig.~\ref{fig:cosine_similarity} shows the cosine similarity between online prototypes and global prototypes (prototypes of the entire memory bank) at each time step.
For the first mini-batch of each task, online prototypes are equal to global prototypes (similarity is 1, omitted in Fig.~\ref{fig:cosine_similarity}).
In the first task, online and global prototypes are updated synchronously with the model updates, resulting in high similarity. 
In subsequent tasks, the model initially learns inadequate features of new classes, causing online prototypes to be inconsistent with global prototypes and low similarity, which shows that accumulating early features as prototypes may be harmful to new tasks. However, the similarity will improve as the model learns, because the model gradually learns representative features of new classes.
Furthermore, the similarity on old classes is only slightly lower, showing that online prototypes are resistant to forgetting. 

\section{Conclusion}

This paper identifies shortcut learning as the key limiting factor for online CL, where the learned features are biased and not generalizable to new tasks. It also sheds light on why the online learning models fail to generalize well.
Based on these,
we present a novel online prototype learning (OnPro) framework to address shortcut learning and mitigate catastrophic forgetting.
Specifically,
by taking full advantage of introduced online prototypes,
the proposed \methodname aims to learn representative features of each class and discriminative features between classes for achieving an equilibrium status that separates all seen classes well when learning new classes, 
while the proposed \dataaugname is able to sense easily misclassified classes and enhance their decision boundaries with the feedback of online prototypes.
Extensive experimental results on widely-used benchmark datasets validate the effectiveness of the proposed \frameworkName as well as its components.
In the future, we will try more efficient alternatives, such as designing a margin loss to ensure discrimination between classes further.

{\small
\bibliographystyle{ieee_fullname}

}

\clearpage

\renewcommand{\thetable}{A\arabic{table}}
\renewcommand{\thefigure}{A\arabic{figure}}
\renewcommand{\theequation}{A\arabic{equation}}
\setcounter{figure}{0}
\setcounter{table}{0}  
\setcounter{equation}{0}  

\appendix
\section*{Appendix}
\section{Difference from PCL}
\label{appendix:PCL}
PCL~\cite{PCL} bridges instance-level contrastive learning with clustering based on unsupervised representation learning. We discuss the differences between PCL and \methodname in the following three parts.

\paragraph{(1) Difference in learning settings.}
PCL is an unsupervised contrastive learning method while \methodname explicitly leverages class labels to compute online prototypes. Thus, \methodname belongs to the supervised setting.

\paragraph{(2) Difference in prototype calculation.}
At each time step (iteration), PCL uses all samples of classes to obtain prototypes by performing K-means clustering.
In contrast, \methodname just utilizes a mini-batch of training data to calculate online prototypes.

\paragraph{(3) Difference in contrastive form (most significant differences).}
The anchor of \methodname as well as its positive and negative samples are online prototypes, which means no instance is involved, while PCL takes instance-level representation as the anchor and cluster centers as the positive and negative samples. 
Specifically,
\methodname regards an online prototype and its augmented view as a positive pair; online prototypes of different classes are regarded as negative pairs. 
PCL clusters samples $M$ times, then regards a representation $\mathbf{z}$ of one image (instance) and its cluster center $\mathbf{c}$ as a positive pair; $\mathbf{z}$ and other cluster centers as negative pairs, formally defined as:
\begin{align}
\label{eq:PCL_infoNCE}
    \mathcal{L}_{\mathrm{PCL}}
    \!=-\sum_{i=1}^{2N}\left( \!\frac{1}{M} \sum_{m=1}^M\!\log\! 
    \frac
    {\exp (\frac{\mathbf{z}_i^\mathrm{T} \mathbf{c}_i^m}{\tau^m})}
    {\sum_{j=0}^{r} 
    \exp (\frac{\mathbf{z}_i^\mathrm{T} \mathbf{c}_j^m}{\tau^m})}\right),
\end{align}
where $N$ is the batch size, $r$ is the number of negative samples, and $\tau^m$ is the temperature hyper-parameter.

In addition, at each iteration, PCL needs to cluster all samples $M$ times, which is very expensive for training, while our \methodname only needs to compute online prototypes once.

\section{Extra Experimental Results}
\label{appendix:extra_exp}
\subsection{More Visual Explanations}
\label{appendix:more_visual}
\begin{figure*}[ht]
  \centering
  \includegraphics[width=1.0\linewidth]{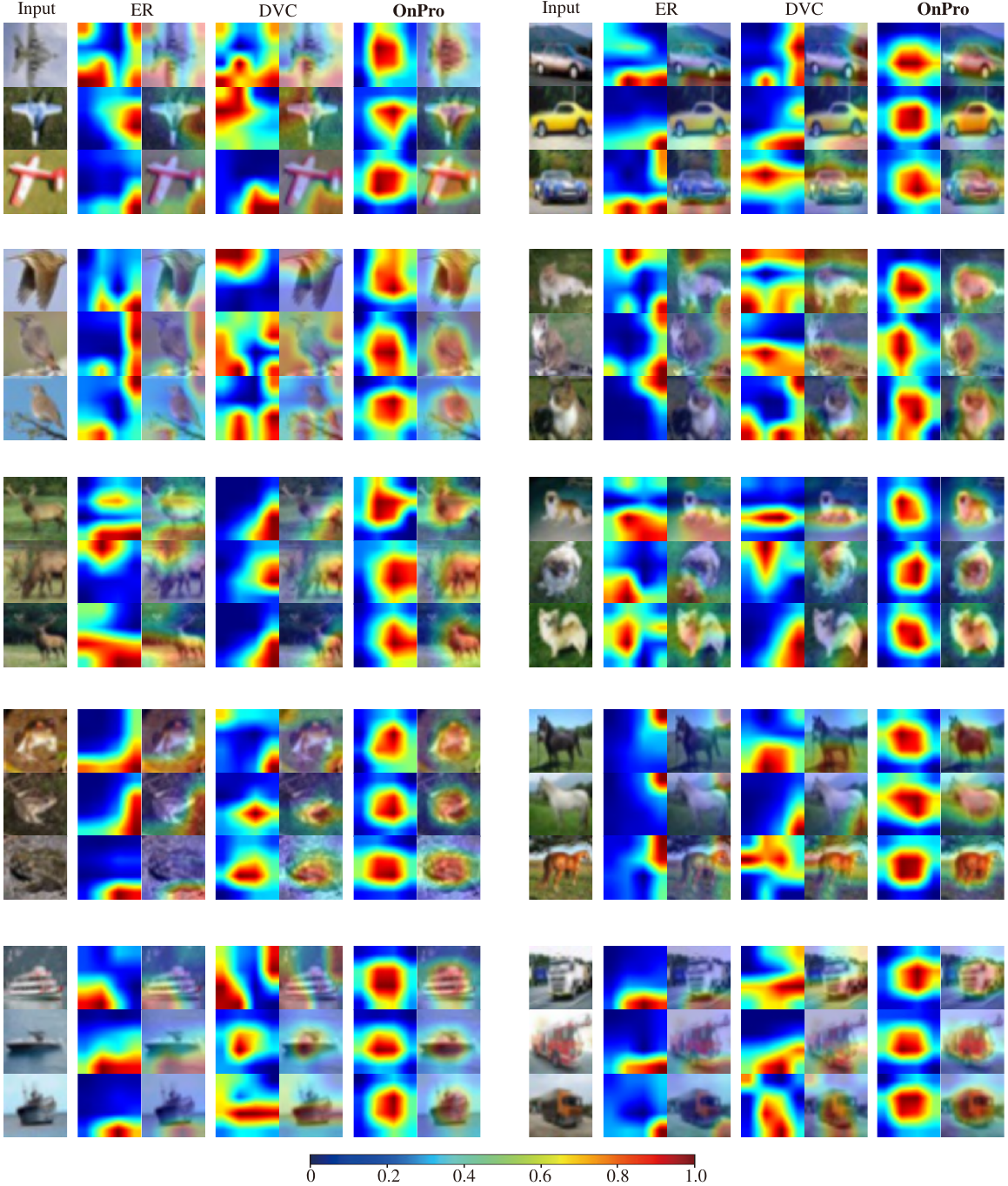}
  \caption{More visual explanations by GradCAM++ on the training set of CIFAR-10 (image size 32 $\times$ 32).}
  \label{fig:moreVisual}
  \vspace{5mm}
\end{figure*}

To further demonstrate the shortcut learning in online CL, we randomly select several images from all (ten) classes in the training set of CIFAR-10 and provide their visual explanations by GradCAM++~\cite{Grad-cam++}, as shown in Fig.~\ref{fig:moreVisual}.
The results confirm that shortcut learning is widespread in online CL. Although ER~\cite{ER} and DVC~\cite{DVC} predict the correct class, they still focus on some oversimplified and object-unrelated features. In contrast, our \frameworkName learns representative features of classes.

\subsection{Knowledge Distillation on ER}
As analyzed in the main paper, it is hard to distill useful knowledge due to shortcut learning. To demonstrate this, we apply the knowledge distillation in \cite{protoAug} to ER, and the results are shown in Table~\ref{tab:KD_ER}.
The performance of ER decreases after using knowledge distillation, and a larger memory bank does not result in significant performance gains.
\begin{table}[h]
\small
\begin{center}
\begin{tabular}{c|rrr}
\shline
\multicolumn{1}{c|}{Method}       & ${M=0.1k}$   & ${M=0.2k}$   & ${M=0.5k}$     \\ \hline
ER & 19.4\std{$\pm$0.6} & 20.9\std{$\pm$0.9} & 26.0\std{$\pm$1.2} \\
ER with KD & 17.0\std{$\pm$2.7} & 17.3\std{$\pm$2.1} & 17.6\std{$\pm$0.8} \\ 
\shline
\end{tabular}
\end{center}
\caption{Average Accuracy with knowledge distillation~\cite{protoAug} (KD) for ER on CIFAR-10. All results are the average of 5 runs.}
\label{tab:KD_ER}
\end{table}

\subsection{Experiments on Larger Datasets}
We conduct extra experiments on ImageNet-100 and ImageNet-1k. ImageNet-100
is a subset of ImageNet-1k with randomly sampled 100 classes; we follow~\cite{Imagenet_spilt} to use the fixed random seed (1993) for dataset generation. We set the number of tasks to 50, the batch size and the buffer batch size to 10, and the memory bank size to 1k for ImageNet-100 and 5k for ImageNet-1k. For a fair comparison, all methods use the same data augmentations, including resized-crop, horizontal-flip, and gray-scale. 
The mean Average Accuracy over 3 runs are reported in Table~\ref{tab:imagenet100_1k}, suggesting: (i) on larger datasets, our OnPro still achieves the best performance and is more stable (lower STD); and (ii) the performance on larger datasets varies greatly. For example, on ImageNet-1k, DVC fails, ER is unstable (large STD), and SCR performs even worse than ER.
\begin{table}[h]
\small
\begin{center}
\resizebox{\columnwidth}{!}{
\begin{tabular}{l|ccccc}
\shline
& ER & SCR & DVC & OCM & \textbf{OnPro}     \\ \hline
IN-100 & 9.6\std{$\pm$3.5} & 12.9\std{$\pm$2.2} & 11.7\std{$\pm$2.9} & 16.4\std{$\pm$3.6} & \textbf{18.6}\std{$\pm$2.3} \\
IN-1k & 5.6\std{$\pm$4.5} & 4.7\std{$\pm$0.2} & 0.1\std{$\pm$0.1} & 5.5\std{$\pm$0.1} & \textbf{6.0}\std{$\pm$0.2} \\
\shline
\end{tabular}
}
\end{center}
\caption{Average Accuracy on ImageNet-100 ($M=1k$) and ImageNet-1k ($M=5k$). 
All results are the average of 3 runs.
}
\label{tab:imagenet100_1k}
\end{table}

\subsection{Visualization of All Classes} 
\begin{figure*}[ht]
  \centering
  \includegraphics[width=1.0\linewidth]{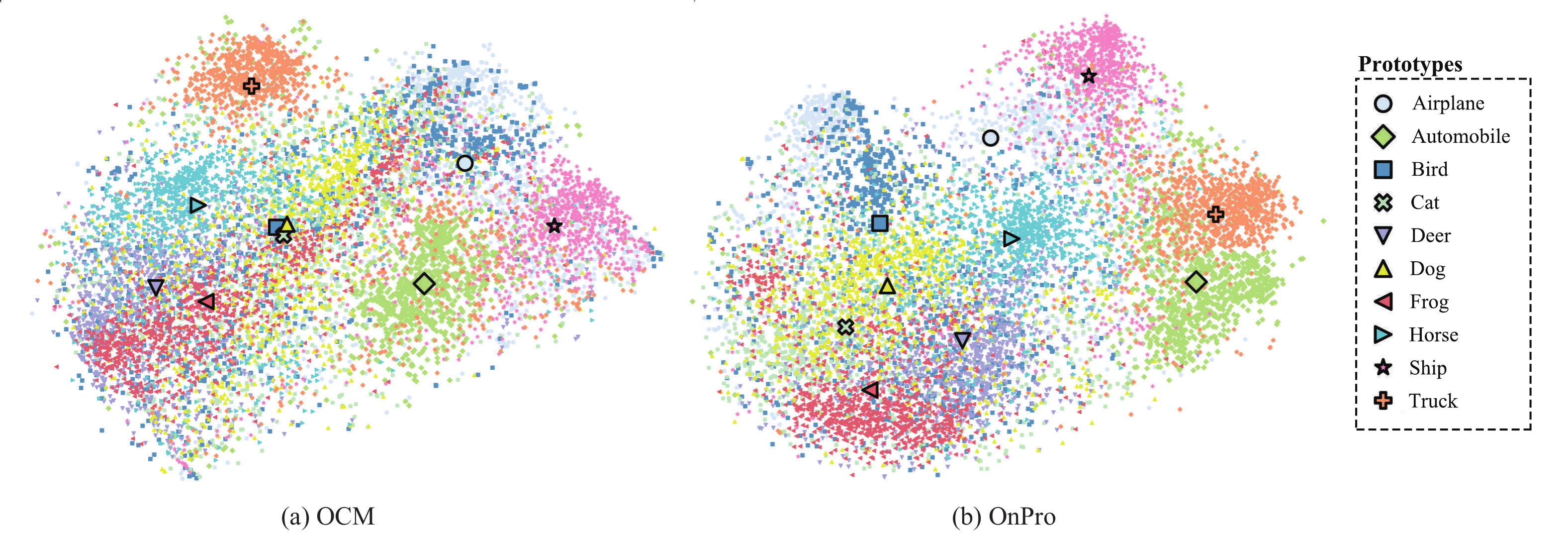}
  \caption{$t$-SNE visualization of all classes in the {test set} of CIFAR-10 ($M=0.2k$).
  }
  \label{fig:visual_cifar10_all}
\end{figure*}
\begin{table*}[ht]
\small
\begin{center}
\begin{tabular}{ccccc}
\shline
\multirow{2}{*}{{Method}} & \multicolumn{2}{c}{CIFAR-10} & \multicolumn{2}{c}{CIFAR-100} \\
\cline{2-3}\cline{4-5}
& Accuracy $\uparrow$ & Forgetting $\downarrow$ & Accuracy $\uparrow$ & Forgetting $\downarrow$ \\ \midrule 
$\mathcal{L}_\mathrm{CE}(\mathrm{both})$ & 48.5\std{$\pm$2.2} & 46.6\std{$\pm$2.4} & 20.4\std{$\pm$0.6} & 41.0\std{$\pm$0.6} \\
$\mathcal{L}_\mathrm{CE}(\mathrm{sepa})$ & 53.2\std{$\pm$2.1} & 38.9\std{$\pm$2.3} & 18.8\std{$\pm$0.6} & 48.1\std{$\pm$0.8} \\ 
\hline
{\frameworkName} (\textbf{ours}) & \textbf{57.8}\std{$\pm$1.1}& \textbf{23.2}\std{$\pm$1.3} & \textbf{22.7}\std{$\pm$0.7} & \textbf{15.0}\std{$\pm$0.8} \\ 
\shline 
\end{tabular}
\end{center}
\caption{Ablation studies about $\mathcal{L}_\mathrm{CE}$ on CIFAR-10 ($M=0.1k$) and CIFAR-100 ($M=0.5k$). 
$\mathcal{L}_\mathrm{CE}(\mathrm{both})$ means calculating $X$ and $X^\mathrm{b}$ in one CE loss, while $\mathcal{L}_\mathrm{CE}(\mathrm{sepa})$ is calculating $X$ and $X^\mathrm{b}$ separately in two CE losses.
All results are the average of 15 runs.
}
\label{tab:ablation_CE}
\end{table*}
To demonstrate the impact of our \frameworkName on classification, we provide the visualization of \frameworkName and OCM for all classes in the {test set} on CIFAR-10 ($M=0.2k$), as shown in Fig.~\ref{fig:visual_cifar10_all}. It is intuitive that the closer the prototypes of the two classes are, the more confused these two classes become. Obviously, OCM does not avoid class confusion, especially for the three animal classes of \texttt{Bird}, \texttt{Cat}, and \texttt{Dog}, while \frameworkName achieves clear inter-class dispersion. 
Furthermore, compared to OCM, \frameworkName can perceive semantically similar classes and present their relationships in the embedding space. Specifically, for the two classes of \texttt{Automobile} and \texttt{Truck}, their distributions are adjacent in \frameworkName because they have more similar semantics compared to other classes. 
However, OCM cannot capture the semantics relationships, causing the two classes to be relatively far apart. 
The results suggest that \frameworkName can achieve an equilibrium status that separates all
seen classes well by learning representative and discriminative features with online prototypes.

\section{Extra Ablation Studies}
\label{appendix:ablations}
\subsection{Class Balance on Cross-Entropy Loss} 
\label{appendix:ablation_CE}
In Table~\ref{tab:ablation_CE}, we find that the way to calculate the cross-entropy (CE) loss can significantly affect the performance of \frameworkName, where $\mathcal{L}_\mathrm{CE}(\mathrm{both}) = l(y \cup y^\mathrm{b}, \varphi(f(x \cup x^\mathrm{b})))$ and $\mathcal{L}_\mathrm{CE}(\mathrm{sepa}) = l(y, \varphi(f(x))) + l(y^\mathrm{b}, \varphi(f(x^\mathrm{b})))$. Here we omit $\operatorname{aug}$ for simplicity. Both $\mathcal{L}_\mathrm{CE}(\mathrm{both})$ and $\mathcal{L}_\mathrm{CE}(\mathrm{sepa})$ degrade the performance because adding the data of new classes will bring serious class imbalance, causing the classifier to easily overfit to new classes and forget previous knowledge.

\subsection{Effects of Rotation Augmentation} 
As mentioned in the main paper, besides resized-crop, horizontal-flip, and gray-scale, OCM and \frameworkName use Rotation augmentation (Rot) like~\cite{protoAug}. To explore the effects of Rot, we employ it for some SOTA baselines, as shown in Table~\ref{tab:dataAug}. We find that using Rot can improve the performance of baselines except for SCR. However, they are still inferior to \frameworkName.
\begin{table}[ht]
\small
\begin{center}
\begin{tabular}{c|rrr}
\shline
\multicolumn{1}{c|}{Method}       & ${M=0.1k}$   & ${M=0.2k}$   & ${M=0.5k}$     \\ \hline
ER-Rot & 30.1\std{$\pm$1.9} & 34.1\std{$\pm$3.0} & 42.8\std{$\pm$4.1} \\
ASER-Rot & 30.7\std{$\pm$3.5} & 35.8\std{$\pm$0.8} & 43.8\std{$\pm$2.1} \\
SCR-Rot & 35.8\std{$\pm$3.3} & 46.4\std{$\pm$2.4} & 59.8\std{$\pm$2.6} \\
DVC-Rot & 45.3\std{$\pm$4.3} & 58.5\std{$\pm$2.8} & 66.7\std{$\pm$2.1} \\
OCM & 47.5\std{$\pm$1.7} & 59.6\std{$\pm$0.4} & 70.1\std{$\pm$1.5} \\
\frameworkName (\textbf{ours})  & \textbf{57.8}\std{$\pm$1.1} & \textbf{65.5}\std{$\pm$1.0} & \textbf{72.6}\std{$\pm$0.8} \\ 
\shline
\end{tabular}
\end{center}
\caption{Average Accuracy using Rotation augmentation (Rot) on CIFAR-10. All results are the average of 5 runs.}
\label{tab:dataAug}
\end{table}

\begin{table*}[h]
\small
\begin{center}
\begin{tabular}{l|cccccc}
\shline
\multicolumn{1}{c|}{$\alpha$} & 0 & 0.10 & 0.25 & 0.50 & 0.75 & 0.9 \\ \hline
CIFAR-10 & 62.9\std{$\pm$2.5} &63.2\std{$\pm$2.0}  & \textbf{65.5}\std{$\pm$1.0} &65.4\std{$\pm$2.7}  &64.6\std{$\pm$1.8}  &64.1\std{$\pm$2.0}  \\
CIFAR-100 & 22.0\std{$\pm$1.5} &\textbf{22.7}\std{$\pm$0.7} &22.1\std{$\pm$1.1}  &21.7\std{$\pm$1.2}  &21.3\std{$\pm$1.3}  &21.1\std{$\pm$1.1} \\
\shline
\end{tabular}
\end{center}
\caption{Effects of the \dataaugname ratio $\alpha$ on CIFAR-10 ($M=0.2k$) and CIFAR-100 ($M=0.5k$).
All results are the average of 5 runs.
}\label{tab:alpha}
\end{table*}

\subsection{Effects of the \dataaugname Ratio $\alpha$}
Encouraging the model to have a tendency to focus on confused classes is helpful for mitigating catastrophic forgetting. However, excessive focus on these classes may disrupt the established equilibrium. 
Therefore, we study the trade-off factor $\alpha$ on CIFAR-10 (${M=0.2k}$) and CIFAR-100 (${M=0.5k}$), and the results are shown in Table~\ref{tab:alpha}.
On the one hand, when $\alpha$ is too small, the \dataaugname reduces to the random selection and takes little account of easily misclassified classes.
On the other hand, too large $\alpha$ causes focusing too much on confused classes and ignoring general cases. Based on the experimental results, we set $\alpha=0.25$ on CIFAR-10 and $\alpha=0.1$ on CIFAR-100 and TinyImageNet.

\subsection{Effects of Projection Head $g$}
We employ a projection head $g$ to get representations, which is widely-used in contrastive learning~\cite{SimCLR}. For baselines, SCR~\cite{SCR}, DVC~\cite{DVC}, and OCM~\cite{OCM} also use a projection head to get representations. To explore the effects of the projector $g$ in \frameworkName, we conduct the experiment in Table~\ref{tab:noProjector}. 
The result shows that projector $g$ can only bring a slight performance improvement, and also illustrates that the performance of \frameworkName comes mainly from our proposed components.
\begin{table}[h]
\small
\begin{center}
\begin{tabular}{c|rrr}
\shline
\multicolumn{1}{c|}{Method}       & ${M=0.1k}$   & ${M=0.2k}$   & ${M=0.5k}$     \\ \hline
no Projector & 56.1\std{$\pm$4.7} & 63.3\std{$\pm$1.9} & 71.0\std{$\pm$1.5} \\
\frameworkName (\textbf{ours})  & \textbf{57.8}\std{$\pm$1.1} & \textbf{65.5}\std{$\pm$1.0} & \textbf{72.6}\std{$\pm$0.8} \\ 
\shline
\end{tabular}
\end{center}
\caption{Average Accuracy without projector $g$ on CIFAR-10. All results are the average of 5 runs.}
\label{tab:noProjector}
\end{table}

\subsection{Effects of Memory Bank Batch Size $m$}
\begin{figure}[ht]
  \centering
  \includegraphics[width=1\linewidth]{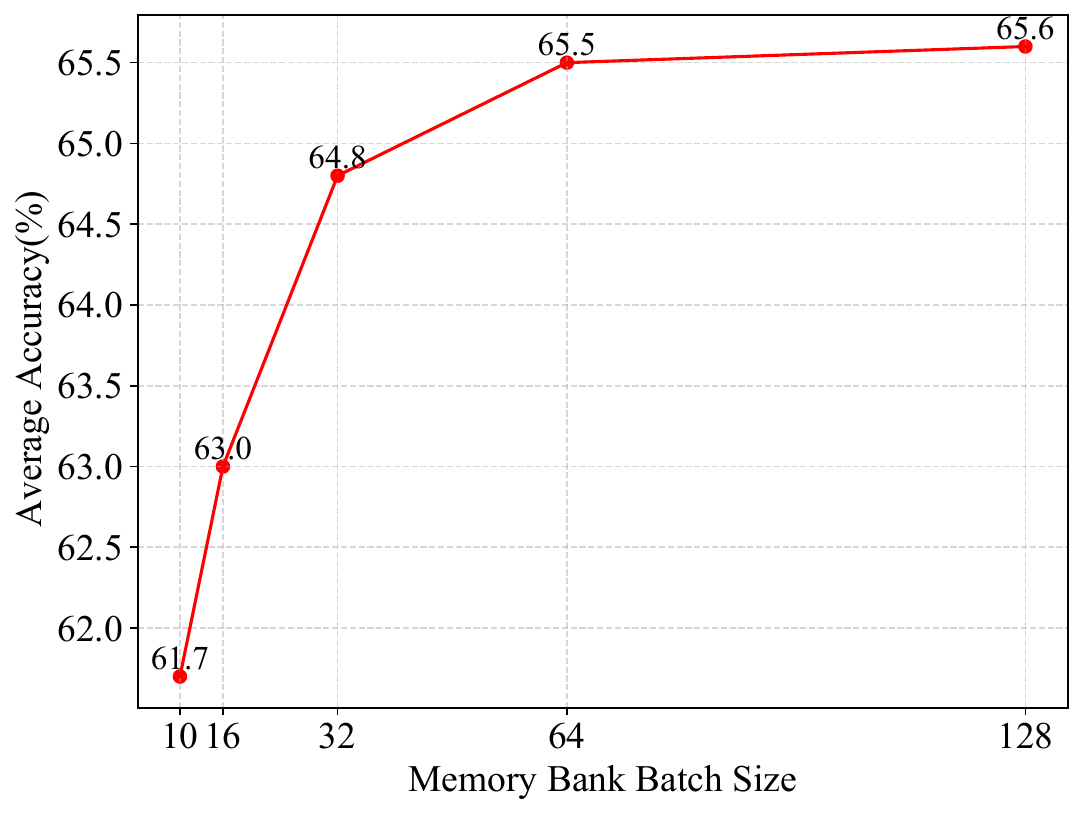}
  \caption{The performance of \frameworkName on CIFAR-10 ($M=0.2k$) with different memory bank batch sizes.}
  \label{fig:memoryBankBatchSize}
\end{figure}

Fig.~\ref{fig:memoryBankBatchSize} shows the effects of memory bank batch size. We can observe that the performance of \frameworkName improves as the memory bank batch size increases. 
However, the training time also grows with larger memory bank batch sizes. 
Following~\cite{OCM}, we set the memory bank batch size to 64.

\section{Training Algorithms of \frameworkName and \dataaugname}
\label{appendix:algorithm}
The training procedures of the proposed \frameworkName and \dataaugname are presented in Algorithms~\ref{alg2} and~\ref{alg_aug}, respectively. The source code will be made publicly available upon the acceptance of this work.
\begin{algorithm*}
    \caption{Training Algorithm of \frameworkName}
    \label{alg2}
    \textbf{Input:}\hspace{0mm} Data stream $\mathfrak{D}$, encoder $f$, projector $g$, classifier $\varphi$, and data augmentation $\operatorname{aug}$.\\
    \textbf{Initialization:} Memory bank $\mathcal{M}$ $\gets$ \{\},\\
    \For{$t$=1 to $T$}{
        \For{each mini-batch $X$ in $\mathcal{D}_t$ }{
            $X^\mathrm{b} \gets$ \dataaugname($\mathcal{M}$)
            \\
            $\widehat{X}, \widehat{X}^\mathrm{b} \gets \operatorname{aug}(X, X^\mathrm{b})$\\
            $\mathbf{z}$, $\mathbf{z^\mathrm{b}}$ = $g$($f$($X \cup \widehat{X}$)), $g$($f$($X^\mathrm{b} \cup \widehat{X}^\mathrm{b}$))\\
            Compute online prototypes $\mathcal{P}$ and $\mathcal{P^\mathrm{b}}$ \Comment{Eq.~\eqref{eq:cal_p} in the main paper}
            \\
            $\mathcal{L}_\mathrm{\frameworkName} \gets$ $\mathcal{L}_\mathrm{\methodname}$($\mathcal{P}$, $\mathcal{P^\mathrm{b}}$) + $\mathcal{L}_{\mathrm{INS}}$($\mathbf{z}, \mathbf{z}^\mathrm{b}$)
            + $\mathcal{L}_\mathrm{CE}$($\varphi(f(\widehat{X}^\mathrm{b}))$)
            \\   
            $\theta_f, \theta_g \gets \mathcal{L}_\mathrm{\frameworkName}$\\
            $\mathcal{M}$ $\gets$ Update($\mathcal{M}, X$) 
        }
    }
\end{algorithm*}
\begin{algorithm*}
    \caption{Algorithm of \dataaugname}
    \label{alg_aug}
    \textbf{Input:}\hspace{0mm} $\mathcal{M}$, and online prototypes $\left\{ \mathbf{p}_i^\mathrm{b} \right\} ^{K^\mathrm{b}}_{i=1}$ of previous time step.\\
    \textbf{Output:}\hspace{0mm} $X^\mathrm{b}$\\
    \textbf{Initialization:} $\mathcal{S}$ $\gets$ \{\}, $n_{\mathrm{\dataaugname}} = \alpha \cdot m$,\\
    $P$ $\gets$Compute probability $P_{i, j}$ for each class pair using $\mathbf{p}_i^\mathrm{b}$ and $\mathbf{p}_j^\mathrm{b}$ \Comment{Eq.~\eqref{eq:cal_d} in the main paper}\\
    \For{each $P_{i,j}$ in $P$}{
        $X_i, X_j$ $\gets$ sample $\lfloor P_{i,j} \cdot n_{\mathrm{\dataaugname}} + 0.5 \rfloor$ images from class $i$ and class $j$\\
        $\mathcal{S}$ $\gets$ $\mathcal{S}$ $\cup$ Mixup($X_i, X_j$)
    }
    $X_{\mathrm{base}} \gets$ the remaining $m-n_{\mathrm{\dataaugname}}$ samples are uniformly randomly selected from $\mathcal{M}$ \\
    $X^\mathrm{b}$ $\gets$ $\mathcal{S}$ $\cup$ Mixup($X_{\mathrm{base}}, X_{\mathrm{base}}$)
    
\end{algorithm*}

\section{Implementation Details about Baselines}
\label{appendix:baselines}
\begin{table*}[ht]
\begin{center}
    \begin{tabular}{l|l}
    \shline
    Baseline & Link\\
    \hline
    iCaRL & \text{https://github.com/srebuffi/iCaRL}\\
    DER++ & \text{https://github.com/aimagelab/mammoth}\\
    PASS & \text{https://github.com/Impression2805/CVPR21\_PASS}\\
    AGEM & \text{https://github.com/facebookresearch/agem}\\
    GSS & \text{https://github.com/rahafaljundi/Gradient-based-Sample-Selection}\\
    MIR & \text{https://github.com/optimass/Maximally\_Interfered\_Retrieval}\\
    GDumb & \text{https://github.com/drimpossible/GDumb}\\
    ASER and SCR & \text{https://github.com/RaptorMai/online-continual-learning}\\
    CoPE & \text{https://github.com/Mattdl/ContinualPrototypeEvolution}\\
    ER and DVC & \text{https://github.com/YananGu/DVC}\\
    OCM & \text{https://github.com/gydpku/OCM}\\
    \shline
    \end{tabular}
\end{center}
\caption{Baselines with source code links.}
\label{tab:baselines}
\end{table*}
The hyperparameters of \frameworkName are given in the main paper. Here we discuss in detail how each baseline is implemented.

For all baselines, we follow their original paper and default settings to set the hyperparameters. We set the random seed to 0 and run the experiment 15 times in the same program to get the results.

For iCaRL, AGEM, and ER, we use the SGD optimizer and set the learning rate to 0.1. We uniformly randomly select samples to update the memory bank and replay.

For DER++, we use the SGD optimizer and set the learning rate to 0.03. We fix $\alpha$ to 0.1 and $\beta$ to 0.5.

For PASS, we use the Adam optimizer and set the learning rate to 0.001. The weight decay is set to 2e-4. We set the loss weights $\lambda$ and $\gamma$ to 10 and fix the temperature as 0.1.

For GSS, we use the SGD optimizer and set the learning rate to 0.1. The number of batches randomly sampled from the memory bank to estimate the maximal gradients cosine similarity score is set to 64, and the random sampling batch size for calculating the score is also set to 64.

For MIR, we use the SGD optimizer and set the learning rate to 0.1. The number of subsamples is set as 100.

For GDumb, we use the SGD optimizer and set the learning rate to 0.1. The value for gradient clipping is set to 10. The minimal learning rate is set to 0.0005, and the epochs to train for the memory bank are 70.

For ASER, we use the SGD optimizer and set the learning rate to 0.1. The number of nearest neighbors to perform ASER is set to 3. We use mean values of Adversarial SV and Cooperative SV, and set the maximum number of samples per class for random sampling to 1.5. We use the SV-based methods for memory update and retrieval as given in the original paper.

For SCR, we use the SGD optimizer and set the learning rate to 0.1. We set the temperature to 0.07 and employ a linear layer with a hidden size of 128 as the projection head.

For CoPE, we use the SGD optimizer and set the learning rate to 0.001. We set the temperature to 1. The momentum of the moving average updates for the prototypes is set to 0.99. We use dynamic buffer allocation instead of a fixed class-based memory as given in the original paper.

For DVC, we use the SGD optimizer and set the learning rate to 0.1. The number of candidate samples for retrieval is set to 50. For CIFAR-100 and TinyImageNet, we set loss weights $\lambda_1 = \lambda_2 = 1$, $\lambda_3 = 4$. For CIFAR-10, $\lambda_1 = \lambda_2 = 1$, $\lambda_3 = 2$.

For OCM, we use the Adam optimizer and set the learning rate to 0.001. The weight decay is set as 0.0001. We set the temperature to 0.07 and employ a linear layer with a hidden size of 128 as the projection head. $\lambda$ is set to 0.5. We set $\alpha$ to 1 and $\beta$ to 2 for contrastive loss and set $\alpha$ to 0 and $\beta$ to 2 for supervised contrastive loss as given in the original paper of OCM.

We refer to the links in Table~\ref{tab:baselines} to reproduce the results.

\section{Execution Time}
\begin{figure*}[ht]
  \centering
  \includegraphics[width=0.8\linewidth]{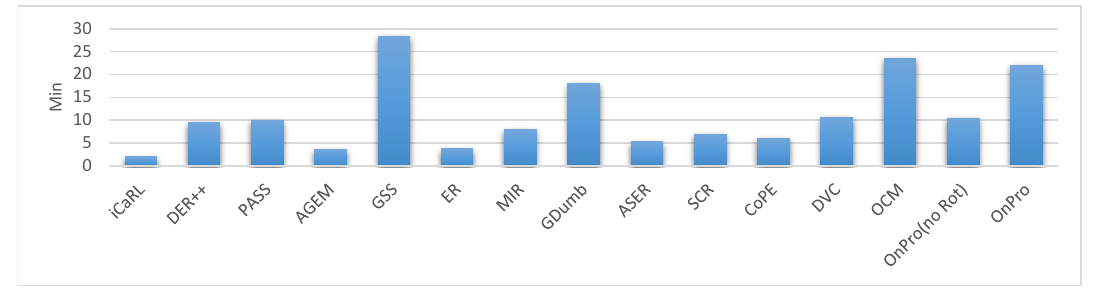}
  \caption{Training time of each method on CIFAR-10.}
  \label{fig:trainingTime}
\end{figure*}
Fig.~\ref{fig:trainingTime} shows the training time of all methods on CIFAR-10. \frameworkName is faster than OCM~\cite{OCM} and GSS~\cite{GSS}. We find that rotation augmentation (Rot) is the main reason for the increase in training time. 
When rotation augmentation is not used, the training time of \frameworkName is significantly reduced and is close to most of the baselines.
Furthermore, \frameworkName achieves the best performance compared to all baselines.

\end{document}